\documentclass[letterpaper]{article} % DO NOT CHANGE THIS
\usepackage[]{aaai2026}  % DO NOT CHANGE THIS
\usepackage{times}  % DO NOT CHANGE THIS
\usepackage{helvet}  % DO NOT CHANGE THIS
\usepackage{courier}  % DO NOT CHANGE THIS
\usepackage[hyphens]{url}  % DO NOT CHANGE THIS
\usepackage{graphicx} % DO NOT CHANGE THIS
\usepackage{natbib}  % DO NOT CHANGE THIS AND DO NOT ADD ANY OPTIONS TO IT
\usepackage{caption} % DO NOT CHANGE THIS AND DO NOT ADD ANY OPTIONS TO IT
\usepackage{xcolor}
\usepackage{orcidlink}
\usepackage{amsmath}
\usepackage{booktabs}
\usepackage{multirow}
\usepackage{amssymb}
\usepackage[linesnumbered,ruled,vlined]{algorithm2e}
\usepackage{todonotes}
\usepackage{mathtools}
\iftrue
\title{Anytime Optimal Decision Tree Learning with Continuous Features}
 \author {
     Harold Silvère Kiossou\orcidlink{0000-0001-6972-9885}\textsuperscript{\rm 1},
     Pierre Schaus\orcidlink{0000-0002-3153-8941}\textsuperscript{\rm 1}
     Siegfried Nijssen\orcidlink{0000-0003-2678-1266}\textsuperscript{\rm 1,2},
}
 \affiliations {
     \textsuperscript{\rm 1}UCLouvain, ICTEAM, Louvain-la-Neuve, Belgium\\
     \textsuperscript{\rm 2}KU LEUVEN, DTAI, Leuven, Belgium \\
     harold.kiossou@uclouvain.be , siegfried.nijssen@kuleuven.be, pierre.schaus@uclouvain.be
 }
\fi

\usepackage{bibentry}
\begin{document}

\maketitle

\begin{abstract}
In recent years, significant progress has been made on algorithms for learning optimal decision trees, primarily in the context of binary features. Extending these methods to continuous features remains substantially more challenging due to the large number of potential splits for each feature.
Recently, an elegant exact algorithm was proposed for learning optimal decision trees with continuous features; however, the rapidly increasing computational time limits its practical applicability to shallow depths (typically 3 or 4). It relies on a depth-first search optimization strategy that fully optimizes the left subtree of each split before exploring the corresponding right subtree. While effective in finding optimal solutions given sufficient time, this strategy can lead to poor anytime behavior: when interrupted early, the best-found tree is often highly unbalanced and suboptimal. In such cases, purely greedy methods such as C4.5 may, paradoxically, yield better solutions.
To address this limitation, we propose an anytime, yet complete approach leveraging limited discrepancy search, distributing the computational effort more evenly across the entire tree structure, and thus ensuring that a high-quality decision tree is available at any interruption point.
Experimental results show that our approach outperforms the existing one in terms of anytime performance.
\end{abstract}

% Uncomment the following to link to your code, datasets, an extended version, or similar.
\begin{links}
    \link{Code}{https://anonymous.4open.science/r/contree-rs-C7B8}
    \link{Datasets}{https://anonymous.4open.science/r/contree-rs-C7B8/datasets/}
\end{links}

\section{Introduction}
Decision trees are widely used in machine learning due to their simplicity and interpretability, with applications in domains such as healthcare, finance, and education. Learning decision trees is typically performed using greedy algorithms, such as CART~\cite{breiman1984cart} and C4.5~\cite{quinlan2014c45}, which construct trees in a top-down manner by selecting a split at each node according to a heuristic criterion. While these greedy approaches are highly scalable, they often produce trees that are less accurate than their optimal counterparts~\cite{van2024optimal}. While learning optimal decision trees (ODTs, trees that minimize the classification error on the training set) is an NP-hard problem, recent advances in computational power and algorithmic techniques have enabled the development of methods that successfully learn such trees. These algorithms leverage combinatorial optimization techniques from mixed-integer linear programming (MILP)~\cite{bertsimas2017optimal,aghaei2021strong}, constraint programming~\cite{verhaeghe2020learning}, and SAT~\cite{narodytska2018learning}, but they struggle to scale with dataset size.
Dynamic programming (DP) and branch-and-bound (BnB) based approaches~\cite{aglin2020learning,demirovic2022murtree} significantly improve scalability, but cannot directly handle numeric features. Consequently, these methods either rely on binarization, which leads to a loss of optimality, or introduce a binary feature for each possible threshold of a numeric attribute. The latter severely harms scalability, as the runtime grows exponentially with the number of features.

To address this limitation, Quant-BnB~\cite{mazumder2022quant} is a specialized algorithm designed to learn optimal decision trees directly from numeric data. It considers splits at selected quantiles of the feature distribution and uses the resulting solutions to prune other parts of the search space. While Quant-BnB significantly outperforms previous approaches on numeric features, it does not scale beyond depth $3$. ConTree~\cite{brița2025optimal} is a subsequent algorithm that combines dynamic programming and branch-and-bound techniques with novel bounding strategies and a specialized solver for depth-2 optimal decision trees. This enables learning depth-4 optimal decision trees on medium-sized numeric datasets within a reasonable time. 

\begin{figure}[t]
\centering
\includegraphics[width=\linewidth]{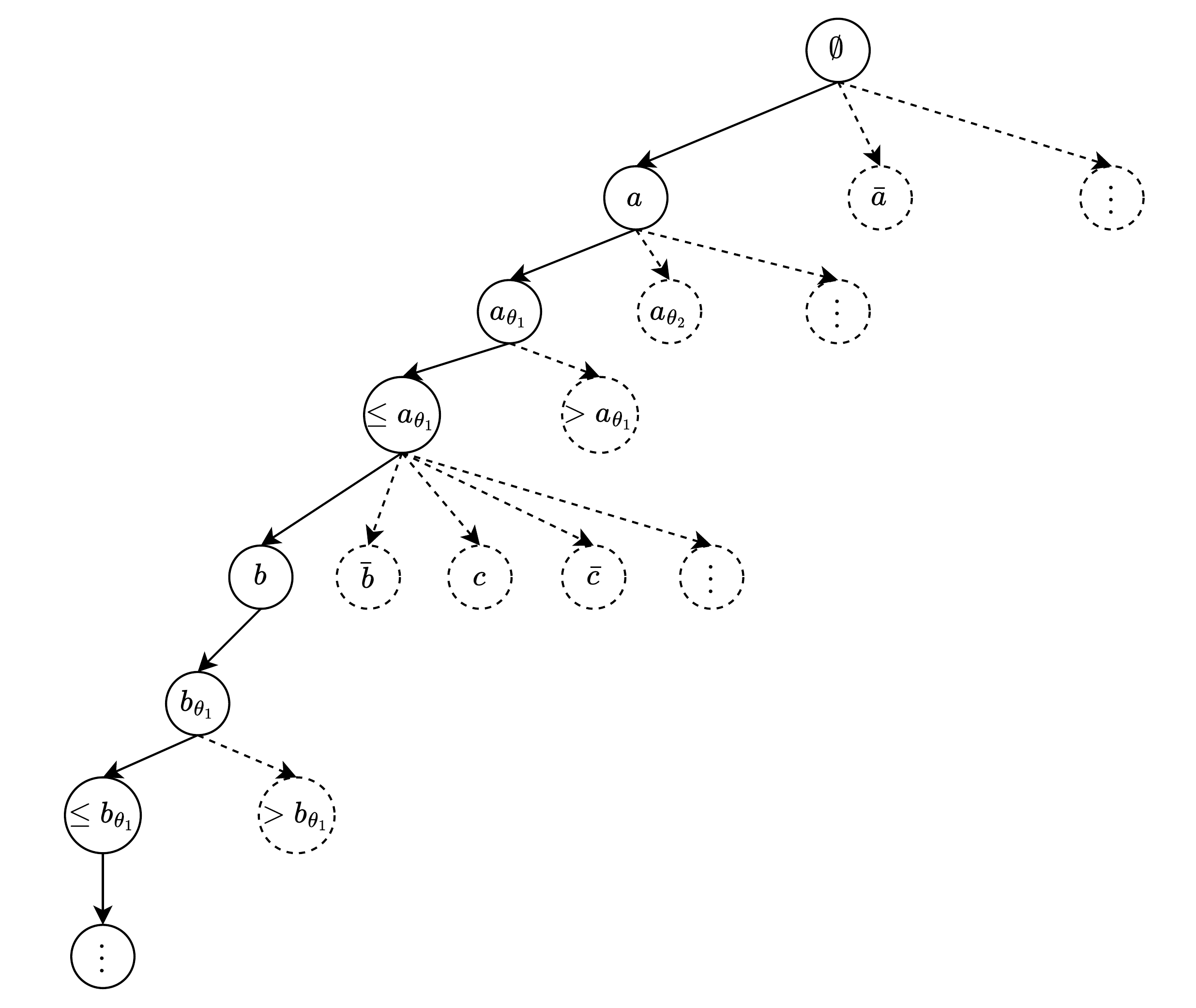}
\caption{ConTree explores the leftmost branches first, often leading to poor anytime performance when interrupted early.}
\label{fig:stuck_search}
\end{figure}

While effective at proving optimality, these methods are not anytime algorithms, as they prioritize bound tightening over early solution quality. Approaches such as DL8.5~\cite{aglin2020learning} or ConTree explore the search space in a depth-first fashion, which often causes it to become stuck in unpromising regions, as illustrated in Figure~\ref{fig:stuck_search}. As a result, it may return poor trees when stopped early. Greedy methods like C4.5 provide quick results but lack the capacity to improve or guarantee optimality over time. Neither approach provides the benefits of a true anytime algorithm, which should quickly produce a good initial solution and then continuously improve it as time permits, eventually proving optimality.

To improve the anytime performance or scalability of exact methods, three notable works have been proposed. LDS-DL8.5~\cite{kiossou2022time} integrates limited discrepancy search (LDS) into DL8.5, resulting in an algorithm that is both anytime and complete. Top-$k$-DL8.5~\cite{blanc2024harnessing} modifies DL8.5 by restricting the candidate features at each node to the Top-$k$ according to a ranking heuristic. This is a compromise between C4.5 and DL8.5: faster and more scalable, but unable to guarantee convergence to the optimal tree.
Finally, the Blossom algorithm~\cite{demirovic2023blossom} follows a fundamentally different search strategy. It uses a depth-first approach that expands decision tree nodes level by level, offering improved anytime behavior by avoiding the possible result of highly unbalanced trees when interrupted early, as for DL8.5. Blossom is guaranteed to find the optimal tree given sufficient time. 

In this work, we introduce CA-ConTree, a novel algorithm for learning decision trees on continuous features that is :
\textbf{complete} (guarantees optimality when given sufficient time),
\textbf{anytime} (produces high-quality solutions early and improves them over time). Limited discrepancy search allows CA-ConTree to prioritize promising regions of the search space early and progressively broaden exploration over time. Our experimental results show that CA-ConTree exhibits strong anytime behavior, outperforming ConTree on medium-sized datasets at depths $4$ and $5$, while producing trees that generalize better under time-limited conditions.

\section{Related Works}
\paragraph{Greedy approaches} 
Learning decision trees has historically relied on greedy algorithms such as CART~\cite{breiman1984cart} and C4.5~\cite{quinlan2014c45}, which optimize local splitting criteria at each node. While these approaches scale well, they are prone to suboptimal solutions due to their myopic nature. In particular, greedy induction tends to produce trees that are larger than the optimal tree on average~\cite{murthy1995decision}, or, when constrained to a fixed depth, yields lower out-of-sample accuracy than optimal trees under the same size limit~\cite{van2024optimal}. As a result, recent research has shifted toward Optimal Decision Trees (ODTs) that leverage combinatorial optimization techniques. These include Mixed-Integer Programming formulations~\cite{bertsimas2017optimal} and logic-based paradigms such as SAT~\cite{narodytska2018learning}, MaxSAT~\cite{hu2020learning}, and Constraint Programming (CP)~\cite{verhaeghe2020learning}. However, a persistent challenge remains: the inherent NP-hardness of optimal decision tree induction severely limits the scalability of these approaches when applied to large datasets or more complex model configurations.

\paragraph{Dynamic programming approaches}
\citet{nijssen2007mining,nijssen2010optimal}, introduced DL8, an early dynamic programming approach for optimal decision tree learning on binary features. This was later improved by~\citet{aglin2020learning} with DL8.5, which incorporates branch-and-bound and enhanced caching techniques to significantly reduce the search space. Other works focused on deriving stronger bounds:~\citet{hu2019optimal} and \citet{lin2020generalized} proposed new lower bounds, including a subproblem similarity bound, to prune redundant computations. More recently, \citet{demirovic2022murtree} introduced a specialized subroutine for depth-two trees and additional constraints to limit the number of branching nodes, further improving efficiency.

\paragraph{Continuous features}
Specialized approaches for learning optimal decision trees (ODTs) directly from continuous data are relatively recent. The first such method, Quant-BnB (Mazumder, Meng, and Wang, 2022), tries to learn optimal decision trees on continuous data by splitting on quantiles of the feature distribution and using the results to bound other parts of the search space. Although it can handle much larger datasets than MIP or SAT-based approaches, it struggles to scale beyond trees of depth three. More recently, \citet{brița2025optimal} proposed ConTree, which leverages the Similarity Lower Bound introduced in OSDT and GOSDT~\cite{hu2019optimal,lin2020generalized} to implement three types of pruning and lower bounds. This allows ConTree to reduce both the computational cost and search space, enabling it to scale better than Quant-BnB and to learn ODTs on medium-sized datasets with a maximum depth of four.

\paragraph{Anytime approaches}
Dynamic programming approaches have one notable limitation: if interrupted, they typically produce an incomplete or unbalanced decision tree, which can be of lower quality than one constructed by greedy heuristics. To address this and improve anytime performance, recent work has explored strategies for generating good intermediate solutions during the search. \citet{kiossou2022time} introduced LDS-DL8.5, which applies iterative Limited Discrepancy Search (LDS) to prioritize solutions that are close to a heuristic baseline tree first. 

Similarly, the Blossom algorithm~\cite{demirovic2023blossom} employs a different search strategy to improve anytime behavior. While it also uses depth-first search, Blossom proceeds layer by layer, always expanding the non-expanded node closest to the root. As with LDS-DL8.5, the first solution found corresponds to the tree that would be produced by a purely greedy strategy.

\section{Background}

\begin{figure*}
    \centering
    \includegraphics[width=\linewidth]{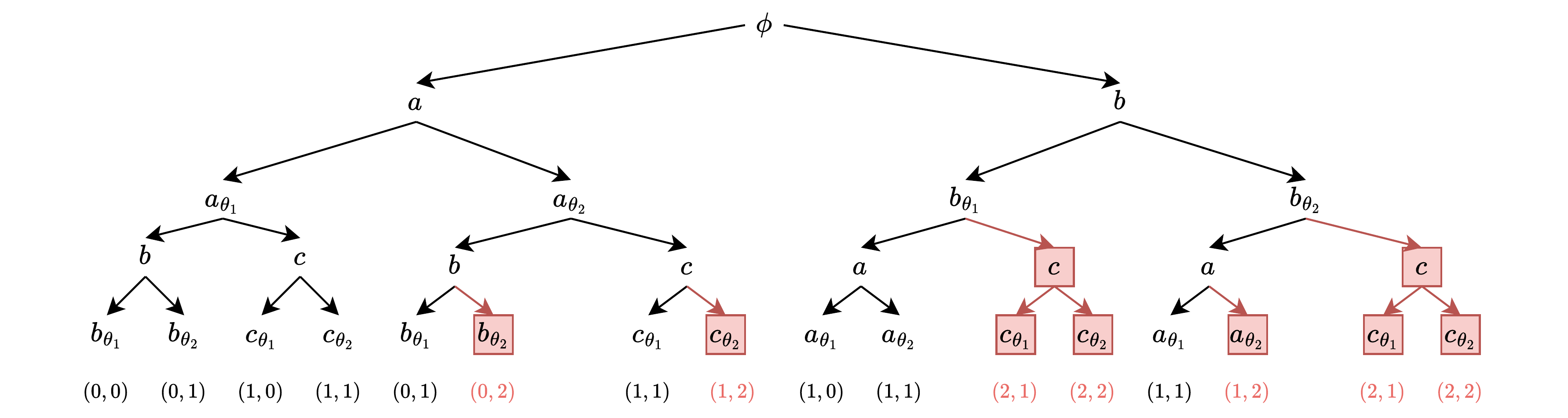}
    \caption{Representation of the search tree for decision tree learning on numeric data with a feature discrepancy budget of $1$ and a split discrepancy budget of $1$. For each feature $f$, candidate split decisions are represented as $f_{\theta_i}$, where $\theta_i$ denotes the index of a possible threshold in the ordered split set $S^f$. Nodes shown in red correspond to decisions that are not explored due to the imposed discrepancy budgets.}
    \label{fig:lds}
\end{figure*}
Learning an optimal decision tree (ODT) is performed on a dataset $\mathcal{D}$ consisting of $n = |\mathcal{D}|$ observations $(x, y)$, where $x \in \mathbb{R}^p$ is the feature vector and $y \in \mathcal{Y}$ is the corresponding label. Let $p$ denote the number of features, with $\mathcal{F} = \{f_1, f_2, \dots, f_p\}$ representing the set of features and $\mathcal{Y}$ the set of classes. For each observation, $x_f$ denotes the value of feature $f$.
For a given feature $f$, let $\mathcal{D}^f$ denote the sorted values of $x_f$ across all $(x, y) \in \mathcal{D}$, and let $\mathcal{U}^f$ denote the set of unique values in $\mathcal{D}^f$. We define the set of candidate split points for feature $f$ as
\begin{equation}
    S^f = \Biggl\{ \frac{\mathcal{U}_1^f + \mathcal{U}_2^f}{2}, \dots, \frac{\mathcal{U}_{|\mathcal{U}^f|-1}^f + \mathcal{U}_{|\mathcal{U}^f|}^f}{2} \Biggl\},
\end{equation}
which contains the midpoints between consecutive unique values of $f$.  Let $m = |S^f|$ be the number of possible thresholds. Given a threshold $\tau \in S^f$, let $\mathcal{D}(f \leq \tau)$ describe the subset of observations $(x, y) \in \mathcal{D}$ where $x_f \leq \tau$.
Let $\mathcal{T}(\mathcal{D}, d)$ describe the set of all decision trees for the dataset $\mathcal{D}$ with a maximum depth of $d$. Then the optimal classification decision tree $t_{opt}$ is the tree that minimizes the misclassification score:

\begin{equation}
    t_{opt} = arg\min_{t \in \mathcal{T}(D, d)} \sum_{(x,y) \in \mathcal{D}} \mathbb{I} \!\left[t(x) \neq y\right].
\end{equation}

\subsection{ConTree}
ConTree (CT) learns ODTs by recursively performing splits on every branching node within a full tree of pre-defined depth. Subproblems are identified by the dataset $\mathcal{D}$ and the remaining depth limit $d$. This can be expressed with the following recurrence equation: 

% \begin{equation}
% CT(\mathcal{D}, d) =
% \begin{cases}
% \displaystyle
% \min_{\hat{y} \in \mathcal{Y}}
% \sum_{(x,y) \in \mathcal{D}}
% \mathbb{I}[\hat{y} \neq y],
% & \text{if } d = 0, \\[1ex]

% \displaystyle
% \min_{f \in \mathcal{F},\, \tau \in S^f}
% \Biggl[
% \begin{aligned}
% & CT(\mathcal{D}(f \leq \tau), d - 1) \\
% & \quad + CT(\mathcal{D}(f > \tau), d - 1)
% \end{aligned}
% \Biggr],
% & \text{if } d > 0.
% \end{cases}
% \end{equation}

\begin{equation}
CT(\mathcal{D}, d) =
\begin{cases}
\displaystyle \min_{\hat{y} \in \mathcal{Y}} \sum_{(x,y) \in \mathcal{D}} \mathbb{I}[\hat{y} \neq y], & \text{if } d = 0 \\[2ex]
\begin{multlined}[b]
    \min_{f \in \mathcal{F}, \tau \in S^f} [ CT(\mathcal{D}_{f \leq \tau}, d - 1) \\ 
    + CT(\mathcal{D}_{f > \tau}, d - 1) ]
\end{multlined}, & \text{if } d > 0.
\end{cases}
\end{equation}
At each internal node, the optimal split is obtained by jointly minimizing over all features $f \in \mathcal{F}$ and their associated thresholds $\tau \in S^f$, selecting the split that minimizes the sum of misclassification scores of the resulting left and right subproblems.
The splits are enumerated dichotomously to progressively narrow the interval of candidate splits. No specific heuristic is used to guide the order of splits.

Since evaluating the misclassification score for all possible thresholds can be computationally expensive, ConTree leverages lower-bound pruning, a specialized depth-two subroutine that iterates over sorted feature data to update class occurrences and efficiently solves, and the same caching mechanisms as in~\cite{demirovic2022murtree} to improve runtime.

% ConTree’s lower-bound pruning techniques rely on the Similarity Lower Bound (SLB) proposed in
% \cite{hu2019optimal,lin2020generalized,demirovic2022murtree}. The SLB assumes that all observations present in the new dataset are classified correctly, while all observations removed from the old dataset were classified incorrectly. This yields the following lower bound:

% \[
% \theta_{\mathcal{D}_{\text{new}}} \ge \theta_{\mathcal{D}_{\text{old}}}
% - \lvert \mathcal{D}_{\text{old}} \setminus \mathcal{D}_{\text{new}} \rvert,
% \]
% where $\theta_{\mathcal{D}}$ is the minimum misclassification score achievable by a decision tree with the same depth limit on dataset $\mathcal{D}$.
% From this bound, several pruning techniques are derived, including \emph{neighborhood pruning}.
% After the misclassification score $\theta_u$ for a split point $u \in [i..j]$ has been computed,
% neighborhood pruning exploits the SLB to discard nearby split points from further consideration.

%\paragraph{Example.}
%Consider a continuous feature with sorted values
%$[0.4, 0.5, 0.5, 0.7, 0.8, 1.0]$, yielding candidate split points
%$[0.45, 0.6, 0.75, 0.9]$.
%Assume that an optimal tree has been computed for the split $\tau = 0.75$ and that its
%misclassification score exceeds the current best upper bound by $\Delta = 2$.
%To obtain an improving solution, a candidate split must therefore reassign at least two
%instances from one side of the split to the other.
%In this example, only the split point $\tau = 0.45$ satisfies this condition, and all other
%split points can be safely pruned.

\section{CA-ConTree}
In this work, we propose the Complete Anytime Continuous Tree (CA-ConTree) algorithm, which leverages limited discrepancy search (LDS)~\cite{harvey1995limited} combined with a split ordering heuristic to build an anytime algorithm for decision tree learning. 
LDS explores solutions in increasing order of deviations from a heuristic, based on the assumption that high-quality solutions differ from the heuristic-guided solution by only a small number of decisions.
This is quite different from standard depth-first search, which blindly explores one branch to its end.
LDS can be implemented on top of depth-first search by enforcing a discrepancy budget during depth-first search and progressively increasing this budget to ensure completeness. At each node, only children within the remaining budget are expanded, and the budget decreases with depth, favoring fewer deviations from the heuristic deeper in the tree.

% As illustrated in Figure~\ref{fig:lds}, at depth $1$ feature $A$ is explored first with a discrepancy budget of $0$. Exploring feature $B$ at the same level requires a budget of at least one, since it deviates from the heuristic choice $A$. The same principle applies at deeper levels, where deviating from the leftmost branch increases the required discrepancy budget to explore alternative nodes.

\begin{algorithm}[!ht]
\SetAlgoLined
\DontPrintSemicolon
%\small
\SetFuncSty{texttt}
\SetKw{Break}{break}
\SetKw{Continue}{continue}
\SetKwProg{Fn}{Procedure}{}{end}
\SetKw{KwOr}{or}
\SetKw{KwAnd}{and}
\SetKwFunction{HasBudget}{budgetRemains}
\SetKwFunction{NextBudget}{nextBudget}
\SetKwFunction{HeuristicSort}{heuristicSort}
\SetKwFunction{Branch}{Branch}

$\theta_{opt} \gets \underset{\hat{y}\in \mathcal{Y}}{\min} \sum_{(x,y)\in\mathcal{D}} \mathbb{I}(\hat{y}\neq y)$

$\mathrm{featDisc} \gets 0, \mathrm{splitDisc} \gets 0$

\While{\HasBudget{$\mathrm{featDisc}, \mathrm{splitDisc}$}}{
\label{alg:main:loop1}
    $\theta_{opt} \gets CT(\mathcal{D}, d, \mathcal{UB}, \mathrm{featDisc}, \mathrm{featDisc})$

    $\mathcal{UB} \gets \min{(\theta_{opt}, \mathcal{UB)}}$
    \label{alg:main:ub}
    
    $(\mathrm{featDisc}, \mathrm{featDisc}) \gets$ \NextBudget{$\mathrm{featDisc}, \mathrm{featDisc}$}
    \label{alg:main:disc_plus}
    
}
\label{alg:main:loop2}

\Fn{$CT(\mathcal{D}, d, \mathcal{UB}, \mathrm{featDisc}, \mathrm{splitDisc})$} {

\lIf{$d = 0$}{
    \Return $\underset{\hat{y}\in \mathcal{Y}}{\min} \sum_{(x,y)\in\mathcal{D}} \mathbb{I}(\hat{y}\neq y)$
}

$\mathcal{F}_{ordered} \gets$ \HeuristicSort{$\mathcal{D}$}
\label{alg:main:heuristic}

\For{$it \gets 0$ \KwTo $|\mathcal{F}_{ordered}| - 1$}{
    $f \gets \mathcal{F}_{ordered}[it]$\;
    
    $disc \gets \mathrm{featDisc} - it$\;
    
    \lIf{$disc < 0$}{
        \Return $\theta_{opt}$
    }

    $\theta_f \gets$ \Branch{$\mathcal{D}, d, f, \mathcal{UB}, disc, \mathrm{splitDisc})$}\;

    \If{$\theta_f < \theta_{opt}$}{
        $\theta_{opt} \gets \theta_f, \mathcal{UB} \gets \theta_{f}$
    }

    % \If{$\theta_{opt} = 0$ \KwOr $\neg \mathtt{timeRemains()}$}{
    %     \Return $\theta_{opt}$
    % }
    
}
\Return $\theta_{opt}$

}

\Return $\theta_{opt}$

\caption{CA-ConTree($\mathcal{D}, d,\mathcal{UB}$)}
\label{algo:main}
\end{algorithm}

Algorithms~\ref{algo:main} and~\ref{algo:branch} present the overall structure of CA-ConTree. Algorithm~\ref{algo:main} describes the main search mechanism. The outer loop (Lines~\ref{alg:main:loop1}–\ref{alg:main:loop2}) repeatedly invokes the subprocedure \textsc{CT} while progressively increasing the discrepancy budgets (Line~\ref{alg:main:disc_plus}) and tightening the global upper bound $\mathcal{UB}$ (Line~\ref{alg:main:ub}).

\begin{algorithm}[ht]
\SetAlgoLined
\DontPrintSemicolon
%\small
\SetKw{Break}{break}
\SetKw{Continue}{continue}
\SetKwFunction{HeuristicSort}{heuristicSort}
%\KwIn{Dataset $\mathcal{D}$, Depth $d$, Feature $f$, SearchConfig $C$, UpperBound $\mathcal{UB}$}
%\KwOut{Best misclassification score $\theta_{opt}$ found for feature $f$}

$\theta_{opt} \gets \underset{\hat{y}\in \mathcal{Y}}{\min} \sum_{(x,y)\in\mathcal{D}} \mathbb{I}(\hat{y}\neq y)$

$splits \gets$ \HeuristicSort{$\{1,2,\dots,m\}$} %\tcp*[r]{Ordered by Gini/Heuristic}
\label{alg:branch:splits}

%$S_{ordered} \gets \text{HeuristicSort}(\text{PossibleSplits}(f))$ \tcp*[r]{Ordered by Gini/Heuristic}
$pruned \gets 0^{|splits|}$ %\tcp*[r]{Bitset of length |splits|, 1 = pruned, 0 = active}

\label{alg:branch:pruned}

\For{$it \gets 0$ \KwTo $|splits| - 1$}{

    % \If{$\neg \mathtt{timeRemains()}$}{\Return $\theta_{opt}$}
    
    $w \gets splits[it]$\;

    $disc \gets \mathrm{splitDisc} - it$\;

    \If{$disc < 0$} {
    \Return $\theta_{opt}$
    }

    \lIf{$pruned[it] = 1$} {
    \Continue
    }

    \lIf{$d = 2$} { $\theta_{w,L}, \theta_{w,R} \leftarrow D2Split(\mathcal{D}, f, w)$}
    \label{alg:branch:d2}
    
    \Else{

        $\mathcal{D}_L \gets \mathcal{D}(f \leq S_w^f), \mathcal{D}_R \gets \mathcal{D}(f > S_w^f)$
        \label{alg:branch:splitdata}
        
        $\theta_{w,L} \leftarrow CT(\mathcal{D}_L, d-1, \mathcal{UB}, \mathrm{featDisc}, disc)$\;

        %$\eta \gets \min\bigl( z(w) - z(i),\; z(j) - z(w) \bigr)$

        $\theta_{w,R} \leftarrow CT(\mathcal{D}_R, d-1, \mathcal{UB} - \theta_{w,L}, \mathrm{featDisc}, disc)$\;
    }

    $\theta_w \leftarrow \theta_{w,L} + \theta_{w,R}$\;
    \label{alg:branch:ub1}

    \If{$\theta_w < \theta_{opt}$}{
        $\theta_{opt} \leftarrow \theta_w$,  $\mathcal{UB} \leftarrow \min(\mathcal{UB}, \theta_w)$\
    }
    \label{alg:branch:ub2}
    
    $\Delta \gets \max(1, \theta_w - \mathcal{UB})$;
    \label{alg:branch:delta1}
    
    \ForEach{$k \in [w - \Delta .. w + \Delta]$}{
    
        $pruned[k] \leftarrow 1$
    }
    \label{alg:branch:delta2}
}
\Return{$\theta_{opt}$}\;
\caption{Branch($\mathcal{D}, d, f, \mathcal{UB}, \mathtt{featDisc}, \mathtt{splitDisc} )$}
\label{algo:branch}
\end{algorithm}

To prevent the search from becoming trapped in deep, unpromising regions of the search space, we introduce two distinct discrepancy budgets: $\mathrm{featDisc}$, which limits the number of features considered at each node, and $\mathrm{splitDisc}$, which limits the number of split thresholds explored for each feature. Since the number of possible thresholds per feature can be large, the search may otherwise spend excessive time exploring deep subtrees without improving the global tree error. By enforcing discrepancy budgets on both features and thresholds, CA-ConTree enables fine-grained control over the growth of the search space and allows exploring different strategies for increasing the budgets to achieve anytime behavior.
Concretely, at a given node, the discrepancy budgets $(\mathrm{featDisc}, \mathrm{splitDisc})$ limit the cumulative number of deviations from the heuristic along a branch: at most $\mathrm{featDisc}$ features can be chosen out of heuristic order, and for each selected feature, at most $\mathrm{splitDisc}$ thresholds can be explored out of the heuristic ranking.
Figure~\ref{fig:lds} illustrates a portion of the search tree for learning a decision tree on numeric data with maximum feature and split discrepancy budgets set to $(1,1)$. The number shown below each branch denotes the cumulative discrepancy cost incurred by the corresponding decision. The leftmost branch corresponds to the best decision according to the heuristic and is therefore expanded first with the minimal budgets $(0,0)$. At depth~1, expanding feature $b$ requires a minimum feature discrepancy of $(1,0)$, while selecting split $\theta_2$ for feature $a$ incurs a split discrepancy of $(0,1)$. Any deviation from the leftmost branch increases the discrepancy budgets, whether the deviation occurs at the feature-selection level or at the split-selection level. As a result, portions of the search space that exceed the available budgets are pruned, as indicated by the red nodes in the figure. With a maximum budget of $(1,1)$, feature $b$ is explored less extensively than feature $a$, since branching on $b$ already exhausts the feature discrepancy budget, and the low split discrepancy budget further limits the exploration of alternative thresholds.

The procedure \textsc{CT} expands features within the current feature discrepancy budget by calling Algorithm~\ref{algo:branch}, processing features in the order determined by a heuristic (Line~\ref{alg:main:heuristic}), which is the Gini score in our implementation. 
Algorithm~\ref{algo:branch} iterates over the candidate split thresholds of a given feature to identify the best split under the current discrepancy budgets. 

In contrast, ContTree \cite{brița2025optimal} enumerates splits in a dichotomic fashion, recursively selecting thresholds at the midpoint of a current interval and using pruning strategies to shrink the interval. However, because it always splits in the middle, this process is not guided by any heuristic. More specifically, let $\{1, 2, \dots,m\}$ denote the indices of the candidate splits in $S^f$ for feature $f$. ConTree first evaluates the split at the midpoint index.

CA-ConTree adopts instead a more classical search strategy. Candidate thresholds are first ordered according to a heuristic measuring their potential impact (Line~\ref{alg:branch:splits}), and the search iterates over them sequentially. For each threshold $w$, Algorithm~\ref{algo:branch} splits the dataset accordingly (Line~\ref{alg:branch:splitdata}) and recursively calls $CT$ at a deeper depth with a reduced split discrepancy budget.
If a split yields a lower objective value than the current best, this value becomes the new best and is used as an updated upper bound for pruning subsequent splits (Lines~\ref{alg:branch:ub1}–\ref{alg:branch:ub2}). When expanding a node whose remaining depth equals two, CA-ConTree invokes the $D2Split$ procedure (Line~\ref{alg:branch:d2}), as proposed in~\cite{brița2025optimal}, to learn an optimal decision tree of depth~2. 
This guarantees that, for each node, the subtree rooted at depth $\mathrm{maxDepth} - 2$ is optimal, improving the overall learning performance. Furthermore, to avoid exploring all possible splits, the search leverages properties derived from the similarity lower bound (SLB), in particular for pruning neighboring candidate thresholds.

\begin{table*}[!ht]
    \centering
    \scriptsize
    \begin{minipage}{0.48\textwidth}
    \centering
        \begin{tabular}{lrrrrrrr}
        \toprule
         & \multicolumn{7}{c}{Runtime (s)} \\
        Approach & 5 & 15 & 30 & 60 & 120 & 300 & 600 \\
        \midrule
        \multirow{1}{*}{CA-ConTree} & \textbf{33.5} & \textbf{26.5} & \textbf{23.1} & \textbf{20.4} & \textbf{18.3} & \textbf{16.3} & \textbf{14.6} \\
        \midrule
        \multirow{1}{*}{ConTree-Gini} & 89.1 & 85.4 & 77.4 & 71.8 & 67.6 & 60.0 & 50.5 \\
        \midrule
        \multirow{1}{*}{ConTree} & 93.8 & 89.0 & 83.7 & 76.8 & 70.3 & 62.1 & 54.2 \\
        \midrule
        \multirow{1}{*}{C4.5} & 40.8 & 40.8 & 40.8 & 40.8 & 40.8 & 40.8 & 40.8 \\
        \bottomrule
\end{tabular}
        \caption{Average primal integral on depth $4$}
        \label{tab:d4_p}
    \end{minipage}
    \hfill
    \begin{minipage}{0.48\textwidth}
    \centering
      \begin{tabular}{lrrrrrrr}
        \toprule
         & \multicolumn{7}{c}{Runtime (s)} \\
        Approach & 5 & 15 & 30 & 60 & 120 & 300 & 600 \\
        \midrule
        \multirow{1}{*}{CA-ConTree} & \textbf{40.2} & \textbf{33.4} & \textbf{31.2} & \textbf{29.5} & \textbf{28.3} & \textbf{26.5} & \textbf{24.8} \\
        \midrule
        \multirow{1}{*}{ConTree-Gini} & 90.1 & 89.2 & 85.2 & 83.2 & 82.2 & 78.4 & 76.7 \\
        \midrule
        \multirow{1}{*}{ConTree} & 93.8 & 93.8 & 93.8 & 93.7 & 92.7 & 83.9 & 79.8 \\
        \midrule
        \multirow{1}{*}{C4.5} & 50.2 & 50.2 & 50.2 & 50.2 & 50.2 & 50.2 & 50.2 \\
        \bottomrule
        \end{tabular}
        \caption{Average primal integral on depth $5$}
        \label{tab:d5_p}
    \end{minipage}

\end{table*}

\paragraph{Similarity Lower Bound and Neighborhood Pruning.} Proposed in~\cite{hu2019optimal, lin2020generalized, demirovic2022murtree}, SLB compares a new dataset $\mathcal{D}_{\text{new}}$ with a previously analyzed dataset $\mathcal{D}_{\text{old}}$ to derive a lower bound on the misclassification score of the new dataset. SLB assumes that all observations present in $\mathcal{D}_{\text{new}}$ are classified correctly, while all observations removed from $\mathcal{D}_{\text{old}}$ are classified incorrectly. This yields the following lower bound:
\begin{equation}
\theta_{\mathcal{D}_{\text{new}}} \ge \theta_{\mathcal{D}_{\text{old}}}
- \lvert \mathcal{D}_{\text{old}} \setminus \mathcal{D}_{\text{new}} \rvert,
\end{equation}
where $\theta_{\mathcal{D}}$ denotes the minimum misclassification score achievable by a decision tree with the same depth limit on dataset $\mathcal{D}$. From this bound, several pruning techniques are derived, including \emph{neighborhood pruning}. After computing the misclassification score $\theta_u$ for a split point $u \in [i..j]$, neighborhood pruning exploits the SLB to discard nearby split points from further consideration.  Algorithm~\ref{algo:branch} leverages this property. When a solution is found for a split $w$ based on the current budgets, the algorithm assumes this solution is the best possible for that split and uses it to prune the neighbor point as done in lines~\ref{alg:branch:delta1} to~\ref{alg:branch:delta2}. Doing so allows to ignore the currently less promising splits. 

\paragraph{Caching.} To avoid redundant computations across search iterations, we employ the same dataset caching strategy as in~\cite{demirovic2022murtree}. This cache is particularly beneficial for calls to the $D2Split$ subroutine. As discrepancy budgets increase, identical subproblems are revisited multiple times; retrieving cached optimal solutions avoids repeated and costly computations. Moreover, within a single search iteration, a cached solution can be reused whenever the current discrepancy budgets are no larger than those associated with the cached entries.

\paragraph{Increasing discrepancy budget.} The discrepancy budgets can be increased following several possible strategies. A simple approach consists of increasing both the feature and split discrepancy budgets linearly, typically by one unit at each iteration. While this steadily enlarges the search space, the potentially large number of candidate splits may delay reaching promising regions, resulting in weak anytime behavior. Alternatively, the budgets can be increased exponentially (doubling each budget at each iteration) to accelerate the exploration of larger portions of the search space. However, this strategy further degrades anytime performance, as rapidly increasing budgets increase the likelihood of becoming trapped in deep and unpromising regions of the search space.
 
\begin{figure}
    \centering
    \includegraphics[width=\linewidth]{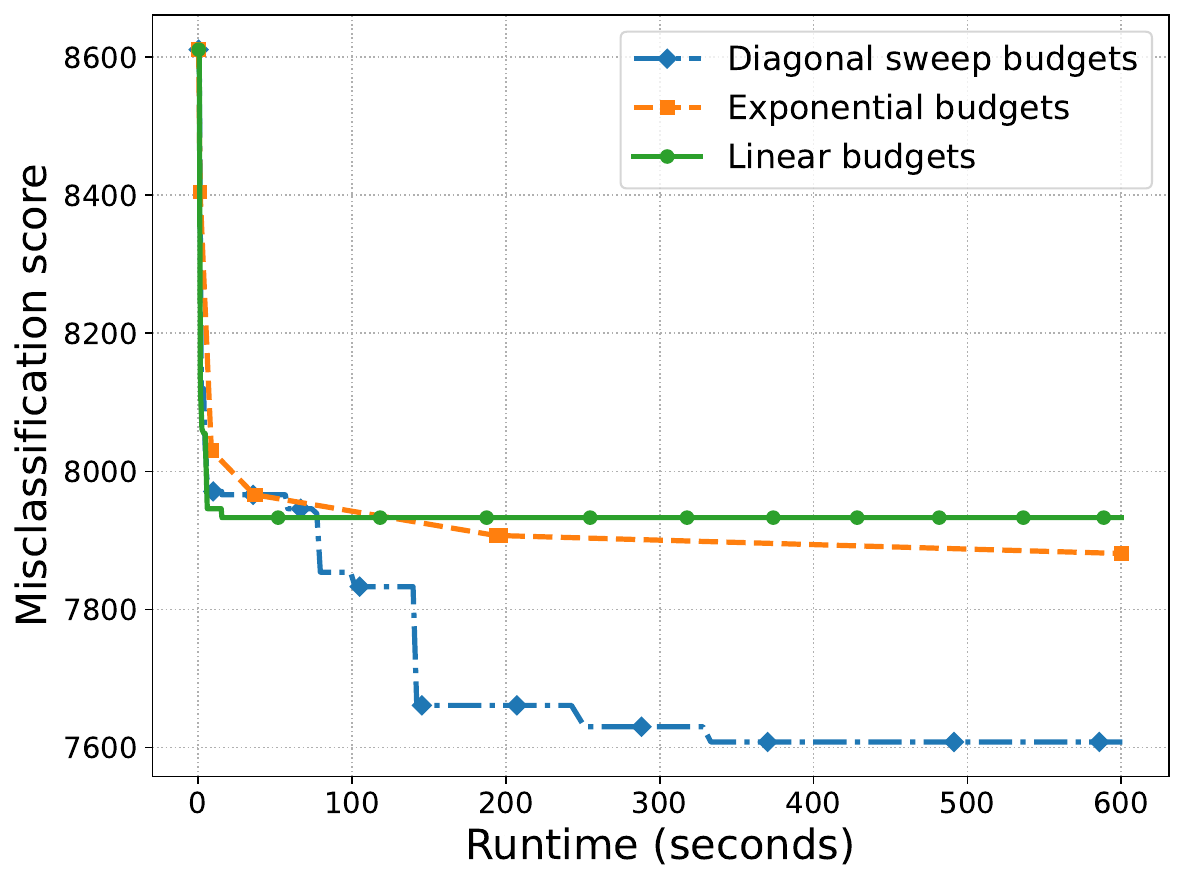}
    \caption{Comparison of discrepancy budget scheduling strategies on a representative dataset (depth 4).}
    \label{fig:discbudget}
\end{figure}

In this work, we adopt a \emph{diagonal sweep} strategy over the two-dimensional discrepancy budget space. Search iterations are ordered according to the sum of the feature and split discrepancy budgets, so that both budgets are increased jointly while keeping their total discrepancy constant. Concretely, the algorithm explores budget pairs in the order
$(0,0)$, $(1,0)$, $(0,1)$, $(2,0)$, $(1,1)$, $(0,2)$, and so on. Figure~\ref{fig:discbudget} shows the evolution of the misclassification score over runtime on a full representative dataset (\textsc{Avila} from the UCI Machine Learning Repository) using CA-ConTree with three different discrepancy budget increase strategies and a timeout of $600$ seconds. The linear strategy plateaus very early, failing to make further progress, while the exponential strategy achieves better solution quality than the linear one. However, the large budget jumps induced by the exponential schedule often cause the search to become trapped in unpromising regions of the search space. In contrast, the diagonal sweep strategy improves solution quality more rapidly and explores a more diverse set of candidate trees, resulting in consistently better anytime performance. Consequently, we adopt the diagonal sweep strategy in all subsequent experiments.

\section{Results}

\begin{table*}[ht]
\centering
\begin{minipage}{0.48\textwidth}
\begin{tabular}{llccc}
\toprule
Dataset & Train Size & C4.5 & Contree & CA-ConTree \\
\midrule
skin & 196045 & 0.984 & \textbf{0.994} & \textbf{0.994} \\
avila & 16693 & 0.620 & 0.616 & \textbf{0.665} \\
occupancy & 16448 & 0.989 & \textbf{0.992} & 0.991 \\
magic & 15216 & 0.820 & 0.732 & \textbf{0.846} \\
htru & 14318 & \textbf{0.978} & 0.935 & 0.977 \\
eeg & 11984 & 0.698 & 0.714 & \textbf{0.765} \\
bean & 10888 & 0.891 & 0.722 & \textbf{0.903} \\
room & 8103 & 0.995 & 0.996 & \textbf{0.997} \\
bidding & 5056 & 0.996 & 0.995 & \textbf{0.997} \\
page & 4378 & 0.964 & 0.963 & \textbf{0.970} \\
wilt & 3871 & 0.978 & 0.980 & \textbf{0.982} \\
rice & 3048 & \textbf{0.919} & 0.917 & 0.916 \\
segment & 1848 & 0.914 & \textbf{0.964} & 0.944 \\
fault & 1552 & 0.667 & 0.677 & \textbf{0.691} \\
bank & 1097 & 0.978 & \textbf{0.986} & 0.985 \\
raisin & 720 & \textbf{0.857} & 0.829 & 0.831 \\
\bottomrule
\end{tabular}
\caption{Average test set accuracy on depth $5$}   
\label{tab:test_d5}
\end{minipage}
\hfill
\begin{minipage}{0.48\textwidth}
\begin{tabular}{llccc}
\toprule
Dataset & Train Size & C4.5 & Contree & CA-ConTree \\
\midrule
skin & 196045 & 0.989 & 0.991 & \textbf{0.997} \\
avila & 16693 & 0.656 & 0.627 & \textbf{0.709} \\
occupancy & 16448 & 0.989 & \textbf{0.992} & 0.991 \\
magic & 15216 & 0.841 & 0.716 & \textbf{0.854} \\
htru & 14318 & \textbf{0.977} & 0.906 & 0.976 \\
eeg & 11984 & 0.722 & 0.711 & \textbf{0.785} \\
bean & 10888 & 0.897 & 0.582 & \textbf{0.908} \\
room & 8103 & \textbf{0.996} & \textbf{0.996} & 0.995 \\
bidding & 5056 & 0.995 & 0.995 & \textbf{0.997} \\
page & 4378 & \textbf{0.967} & 0.957 & \textbf{0.967} \\
wilt & 3871 & \textbf{0.980} & 0.978 & 0.977 \\
rice & 3048 & \textbf{0.915} & 0.890 & 0.900 \\
segment & 1848 & 0.937 & 0.739 & \textbf{0.947} \\
fault & 1552 & 0.673 & 0.626 & \textbf{0.734} \\
bank & 1097 & 0.983 & \textbf{0.988} & 0.980 \\
raisin & 720 & \textbf{0.841} & 0.806 & 0.838 \\
\bottomrule
\end{tabular}

\caption{Average test set accuracy on depth $6$}
\label{tab:test_d6}
\end{minipage}
\end{table*}

We conducted a series of experiments to evaluate the performance of CA-ConTree. 
First, we analyze its anytime behavior in comparison with the baseline ConTree (with and without Gini) approaches, followed by an evaluation of their ability generalize. 
All experiments were performed on an Intel Xeon Platinum 8160 CPU with 320 GB of RAM, running Rocky Linux 8.4. 
The algorithms were evaluated on the 16 datasets from the UCI Machine Learning Repository previously used in ~\cite{brița2025optimal}. 
We compare CA-ConTree to the other algorithms using the \emph{average primal integral}, as introduced in~\cite{BERTHOLD2013611}, to measure the anytime behavior of optimization solvers. The primal integral aims to measure the progress of an algorithm's primal bound convergence toward the optimal (or best known) solution over the entire solving time. It is based on the \textit{primal function} \( p(t) \), which represents the gap between the current solution \( x(t) \) at time \( t \) and the optimal or best known solution \( x_{\text{opt}} \). The \textit{primal gap} of a solution \( x(t) \) is defined as
\[
\gamma(x(t)) = \frac{|x(t) - x_{\text{opt}}|}{|x(t)|}.
\]
The function \( p(t) \) equals 1 if no feasible solution has been found by time \( t \), and \( \gamma(x(t)) \) otherwise. The function \( p(t) \) is a step function that changes whenever a new feasible solution is found. It is monotonically decreasing and becomes zero once the optimal solution is reached. The primal integral \( P(T) \) is defined as the integral of the primal gap function \( p(t) \) over the time horizon \( T \):
\[
P(T) = \int_0^T p(t) \, dt = \sum_{i=1}^{n} p(t_{i-1}) \cdot (t_i - t_{i-1}),
\]
where each \( t_i \) denotes a time point at which a new incumbent solution is found.
The primal integral encourages the early discovery of high-quality solutions. If a better solution is found at the same time, \( P(t_{\max}) \) decreases. Similarly, if the same solution is found earlier, \( P(t_{\max}) \) also decreases. The ratio \( P(t_{\max}) / t_{\max} \) can be interpreted as the average quality of the solution during the search process. A smaller value indicates a higher expected quality of the current solution if the algorithm is interrupted at an arbitrary point in time.

Tables~\ref{tab:d4_p} and~\ref{tab:d5_p} report the evolution of the average primal integral across various time budgets (from $5$ to $600$ seconds) for tree depths $4$ and $5$. We compare CA-ConTree with ConTree using and without the Gini heuristic, as well as with the greedy C4.5 baseline.
Across both depths, CA-ConTree consistently achieves the lowest primal integral for all time budgets, highlighting its strong anytime behavior. In contrast, ConTree and ConTree-Gini improve more slowly and remain substantially behind, while C4.5 yields a constant primal integral due to its non-anytime, greedy nature. The gap between CA-ConTree and the other complete approaches persists even at larger time budgets, indicating that the LDS strategy enables faster discovery of high-quality solutions throughout the search. The primal integral values observed for ConTree and ConTree-Gini indicate a weak anytime behavior. Although these approaches are complete and may eventually reach high-quality or optimal solutions, they tend to maintain poor incumbents for a large portion of the runtime. As a result, improvements in solution quality occur late in the search and contribute only marginally to reducing the accumulated primal gap. This effect is only partially mitigated by the Gini heuristic. C4.5 exhibits constant primal integral values, as it immediately produces a greedy solution and does not refine it further. While C4.5 lacks optimality guarantees, its early availability of a reasonable solution allows it to perform better than plain ConTree in terms of primal integral for small and medium time budgets.

Next, we evaluate the test-set performance of ConTree and CA-ConTree against the baseline C4.5 algorithm using 5-fold cross-validation on the same previous datasets with a timeout of $600$ seconds. Tables~\ref{tab:test_d5} and~\ref{tab:test_d6} report the results for tree depths $5$ and $6$, respectively.

Overall, CA-ConTree achieves the best or near-best test accuracy on the majority of datasets, outperforming both C4.5 and the baseline ConTree in many cases. In particular, ConTree sometimes fails to surpass the C4.5 heuristic, especially on datasets such as \textsc{magic}, \textsc{bean}, and \textsc{htru}, while the improved anytime behavior of CA-ConTree often allows it to close this gap and reach higher-quality solutions within the same time budget.

The benefits of CA-ConTree are most pronounced on medium-sized datasets, where its ability to discover strong solutions early and refine them over time leads to consistent improvements in final performance. On a small number of datasets, C4.5 remains competitive or slightly superior, particularly at larger depths; however, these differences are generally modest. Overall, these results indicate that integrating limited discrepancy search does not degrade generalization performance and instead provides a more robust framework for learning high-quality trees compared to both greedy heuristics and standard exact baselines.

\section{Conclusion}
We introduced CA-ConTree, a novel algorithm for learning optimal decision trees on continuous features with strong anytime performance. By integrating limited discrepancy search and carefully controlling the exploration of both feature and split discrepancies, CA-ConTree can discover high-quality solutions early and improve upon them.
Experimental results on a diverse set of benchmark datasets demonstrated that CA-ConTree consistently outperforms existing exact methods in time-limited settings, particularly on medium-sized datasets and for tree depths where standard approaches struggle. Compared to ConTree, CA-ConTree achieves substantially better anytime behavior without sacrificing generalization performance, and often surpasses both greedy baselines and exact solvers within the same computational budget.
While ConTree remains more effective when the primary objective is to prove optimality on shallow trees, CA-ConTree provides a more practical alternative in scenarios where high-quality solutions are required quickly.
As future work, it could be interesting to combine the two approaches and dynamically switch strategies, for example after a certain amount of time.

\bibliography{aaai2026}


\end{document}